\PassOptionsToPackage{unicode}{hyperref}
\PassOptionsToPackage{hyphens}{url}
\documentclass[
]{article}
\usepackage{xcolor}
\usepackage{amsmath,amssymb}
\usepackage{iftex}
\ifPDFTeX
  \usepackage[T1]{fontenc}
  \usepackage[utf8]{inputenc}
  \usepackage{textcomp} 
\else 
  \usepackage{unicode-math} 
  \defaultfontfeatures{Scale=MatchLowercase}
  \defaultfontfeatures[\rmfamily]{Ligatures=TeX,Scale=1}
\fi
\usepackage{lmodern}
\ifPDFTeX\else
\fi
\IfFileExists{upquote.sty}{\usepackage{upquote}}{}
\IfFileExists{microtype.sty}{
  \usepackage[]{microtype}
  \UseMicrotypeSet[protrusion]{basicmath} 
}{}
\makeatletter
\@ifundefined{KOMAClassName}{
  \IfFileExists{parskip.sty}{%
    \usepackage{parskip}
  }{
    \setlength{\parindent}{0pt}
    \setlength{\parskip}{6pt plus 2pt minus 1pt}}
}{
  \KOMAoptions{parskip=half}}
\makeatother
\usepackage{longtable,booktabs,array}
\usepackage{calc} 
\usepackage{etoolbox}
\makeatletter
\patchcmd\longtable{\par}{\if@noskipsec\mbox{}\fi\par}{}{}
\makeatother
\IfFileExists{footnotehyper.sty}{\usepackage{footnotehyper}}{\usepackage{footnote}}
\makesavenoteenv{longtable}
\usepackage{graphicx}
\makeatletter
\newsavebox\pandoc@box
\newcommand*\pandocbounded[1]{
  \sbox\pandoc@box{#1}%
  \Gscale@div\@tempa{\textheight}{\dimexpr\ht\pandoc@box+\dp\pandoc@box\relax}%
  \Gscale@div\@tempb{\linewidth}{\wd\pandoc@box}%
  \ifdim\@tempb\p@<\@tempa\p@\let\@tempa\@tempb\fi
  \ifdim\@tempa\p@<\p@\scalebox{\@tempa}{\usebox\pandoc@box}%
  \else\usebox{\pandoc@box}%
  \fi%
}
\def\fps@figure{htbp}
\makeatother
\providecommand{\tightlist}{%
  \setlength{\itemsep}{0pt}\setlength{\parskip}{0pt}}
\usepackage{bookmark}
\IfFileExists{xurl.sty}{\usepackage{xurl}}{} 
\hypersetup{
  hidelinks,
  pdfcreator={LaTeX via pandoc}}

\author{}
\date{}

\ifPDFTeX
\usepackage{textcomp}
\DeclareUnicodeCharacter{00B0}{\ensuremath{{}^\circ}}
\DeclareUnicodeCharacter{00D7}{\ensuremath{\times}}
\DeclareUnicodeCharacter{2212}{\ensuremath{-}}
\DeclareUnicodeCharacter{00B7}{\textperiodcentered}
\DeclareUnicodeCharacter{00A7}{\S}
\DeclareUnicodeCharacter{2192}{\ensuremath{\rightarrow}}
\DeclareUnicodeCharacter{00E9}{\'{e}}
\fi

\begin{document}

\section{Perfect Detection, Failed Control: The Geometry of Knowing
vs.~Steering in Language
Models}\label{perfect-detection-failed-control-the-geometry-of-knowing-vs.-steering-in-language-models}

\textbf{Cosimo Galeone} · \texttt{cosimo.galeone@alomana.com}\\
\textbf{Anna Ettorre} · \texttt{anna.ettorre@alomana.com}\\
\textbf{Minsu Park} · \texttt{minsu.park@alomana.com}\\
\textbf{Giuseppe Ettorre} · \texttt{ge@alomana.com}\\
\textbf{Daniele Ligorio} · \texttt{dl@alomana.com}

\emph{Alomana, Grottaglie, Italy}

\begin{center}\rule{0.5\linewidth}{0.5pt}\end{center}

\subsection{Abstract}\label{abstract}

A central aspiration of mechanistic interpretability is
\emph{controllability}: if we understand where a behavior is represented
in a model's activations, we should be able to modify it. This
aspiration rests on a hidden premise --- that the direction which
\emph{detects} a behavior and the direction which \emph{controls} it are
the same, or at least close. We ask whether this is true.

We frame the question geometrically: what is the angle between the
direction that best discriminates a behavior and the direction that best
causes it? If detection implies control, the angle should be near 0°. If
not, the cosine between them quantifies what we call the
\textbf{detection-intervention gap}.

We study two contrasting behaviors on Gemma 2-2B-it. For output format
(clean JSON vs markdown fencing), detection and control collapse onto a
single axis: one and the same direction both classifies the behavior
and, when added, flips it. Hallucination breaks the premise. The model
detects whether an entity is real with perfect linear separability (AUC
= 1.000 from layer 5), yet the direction carrying that signal sits at
\(\cos = 0.12\) --- about 83° --- from the direction that produces a
refusal: a small, reproducible alignment, far from the
\(\cos \approx 1\) that ``detection is control'' would require. A
detector built a second way, from activations with no chosen tokens,
likewise fails to align with control (\(\cos = -0.06\)).

The gap generalizes. Across four models from three families and two
scales (1B--9B), \(\cos\) stays in {[}0.12, 0.20{]}; it is identical
before and after instruction tuning (0.1197 vs 0.1200), so the geometry
is laid down in pretraining. On Gemma 2-2B-it, a 15° rotation from the
detection direction toward the intervention direction partially bridges
the gap --- 73\% and 60\% refusal on two held-out entity categories, at
1.8\% false positives (N=115).

It is tempting to read this cosine as a steerability diagnostic --- high
meaning the detection direction is itself a control knob, low meaning it
is not. We test that reading, and it does not hold. Measured the same
independent way, the cosine sits near the high-dimensional chance level
for steerable and unsteerable behaviors alike; format's apparent
alignment comes only from using one vector in both roles. The reason is
structural: detection is not a single direction but a high-dimensional
class, and what separates the steerable case is \emph{functional} ---
whether the control direction also works as a detector --- not readable
from a static angle. What the cosine \emph{is} is a robust,
weight-computable signature of the dissociation, invariant across four
models --- not a predictor of how steerable a behavior is.

\begin{center}\rule{0.5\linewidth}{0.5pt}\end{center}

\subsection{1. Introduction}\label{introduction}

Recent work in mechanistic interpretability has demonstrated a striking
capability: find a direction in a model's residual stream that
corresponds to a behavior, and you can often steer that behavior by
adding or ablating it. Arditi et al.~(2024) showed that refusal is
mediated by a single linear direction --- the same difference-in-means
direction that \emph{separates} harmful from harmless prompts also
\emph{removes} refusal when ablated. Li et al.~(2023) found
truth-related directions by probing and added them at inference to
increase honesty. Turner et al.~(2023) formalized activation addition as
a general steering paradigm. These results are often read as licensing a
general inference: that the direction which \emph{detects} a behavior is
also the direction that \emph{controls} it.

We test the generality of this paradigm by asking a precise geometric
question: \textbf{what is the angle between the direction that best
detects a behavior and the direction that best controls it?} (Figure 1).

If detection implies control, these directions should be aligned
(\(\cos \approx 1\)). If the relationship is more complex, the angle
between them characterizes the \emph{geometry of the
detection-intervention gap}.

We study this on Gemma 2-2B-it (26 layers, 2304-dimensional residual
stream), using two contrasting behaviors:

\begin{enumerate}
\def\labelenumi{\arabic{enumi}.}
\tightlist
\item
  \textbf{Output format} (markdown fencing vs clean JSON): a binary,
  directly observable rendering choice.
\item
  \textbf{Hallucination} (fabricating answers about non-existent
  entities vs refusing): a behavior entangled with the model's knowledge
  state.
\end{enumerate}

For format the angle is near 0° --- detection is control; for
hallucination it is about 83°, the two directions nearly perpendicular.
The rest of the paper establishes each case, asks why hallucination's
directions diverge, shows the divergence can be partly undone by
rotating the intervention direction, and confirms the pattern across
three model families. One thing this angle does \emph{not} do, we should
say up front, is serve as an a priori predictor of steerability: the
cosine is a signature of the gap, not an oracle for it, and Section 8
makes that negative result precise.

Whether this angle varies systematically across a broader range of
behaviors --- and what predicts where a given behavior falls --- is a
question we leave to future work.

\subsubsection{Contributions}\label{contributions}

\begin{enumerate}
\def\labelenumi{\arabic{enumi}.}
\item
  \textbf{A functional dissociation between knowing and steering.} On
  hallucination, the model detects fake entities with perfect linear
  separability (AUC = 1.000 from layer 5), yet the detection direction
  barely steers behavior, and the direction that \emph{does} steer
  (refusal) is itself only a weak detector. A cross-model double
  dissociation (Section 8) shows detecting and acting are separate
  faculties, not one computation that occasionally fails to connect
  (Sections 4, 8).
\item
  \textbf{The gap is weight-readable but not a steerability oracle.} The
  cosine between detection and intervention is computable from the
  weights and consistent across four models (\(\cos \in [0.12, 0.20]\),
  a small but reproducible positive). It is tempting to read it as an a
  priori steerability test; we show it is not --- detection is a
  high-dimensional class and steerability is a functional property, not
  one readable from a static angle (Section 8). The cosine is a
  signature of the dissociation, not a control dial.
\item
  \textbf{The gap is not a construction artifact.} A detection direction
  built from activations, with no hand-chosen tokens, also fails to
  align with the intervention direction (Section 4.3).
\item
  \textbf{A mechanism for the gap.} We trace it to the model's output
  mapping, where a mechanism that copies the salient entity name
  dominates the epistemic signal --- so steering the detection direction
  makes fabrication worse, not better (Sections 4.3, 5).
\item
  \textbf{A partial fix.} Rotating 15° from the detection direction
  toward the intervention direction partially recovers refusal --- Type
  2 from 13\% to 60\% --- on a held-out stress test (Section 6).
\item
  \textbf{Generality.} The gap holds across four models and two scales
  (1B--9B) and is already present before instruction tuning, placing its
  origin in pretraining (Section 7).
\end{enumerate}

\begin{center}\rule{0.5\linewidth}{0.5pt}\end{center}

\subsection{2. Methods}\label{methods}

All experiments use inference-time analysis with PyTorch forward hooks
on Gemma 2-2B-it (Gemma Team, 2024), float16. No fine-tuning or training
is performed. Cross-model experiments use Llama-3.2-1B-Instruct (Dubey
et al., 2024), Qwen-2.5-1.5B-Instruct (Qwen Team, 2025), Gemma 2-9B-it
(Gemma Team, 2024), and the Gemma 2-2B base model (Gemma Team, 2024).

The central quantity we measure is the cosine between a direction that
\emph{detects} a behavior and the direction that \emph{controls} it. A
direction can be built two ways: \emph{data-driven}, from differences in
activations between two conditions (difference-in-means, Section 2.1),
or \emph{hand-picked}, by reading the \texttt{lm\_head} rows of tokens
typical of a behavior (Section 2.2). We build the \textbf{detection}
direction both ways; the \textbf{intervention} (refusal) direction is
hand-picked from \texttt{lm\_head} alone. We test whether a direction is
causally relevant by adding it to the residual stream and measuring the
behavioral change (Section 2.5), and we localize where in the network
the signal lives (Section 2.3). All analysis is at inference time --- we
observe and steer the model without updating any weights. Figure 1 lays
out the full pipeline.

\begin{figure}
\centering
\includegraphics[width=0.95\linewidth,height=\textheight,keepaspectratio,alt={The method in one view. From the last-token residual state \textbackslash mathbf\{h\} we build two directions: a detection direction --- whether the model internally registers an entity as fake, obtainable either from activations (difference-in-means) or from the output vocabulary (lm\_head) --- and an intervention direction (refusal), read from lm\_head alone. We then measure the cosine between them (the paper's central quantity) and steer by adding a direction to \textbackslash mathbf\{h\} during generation.}]{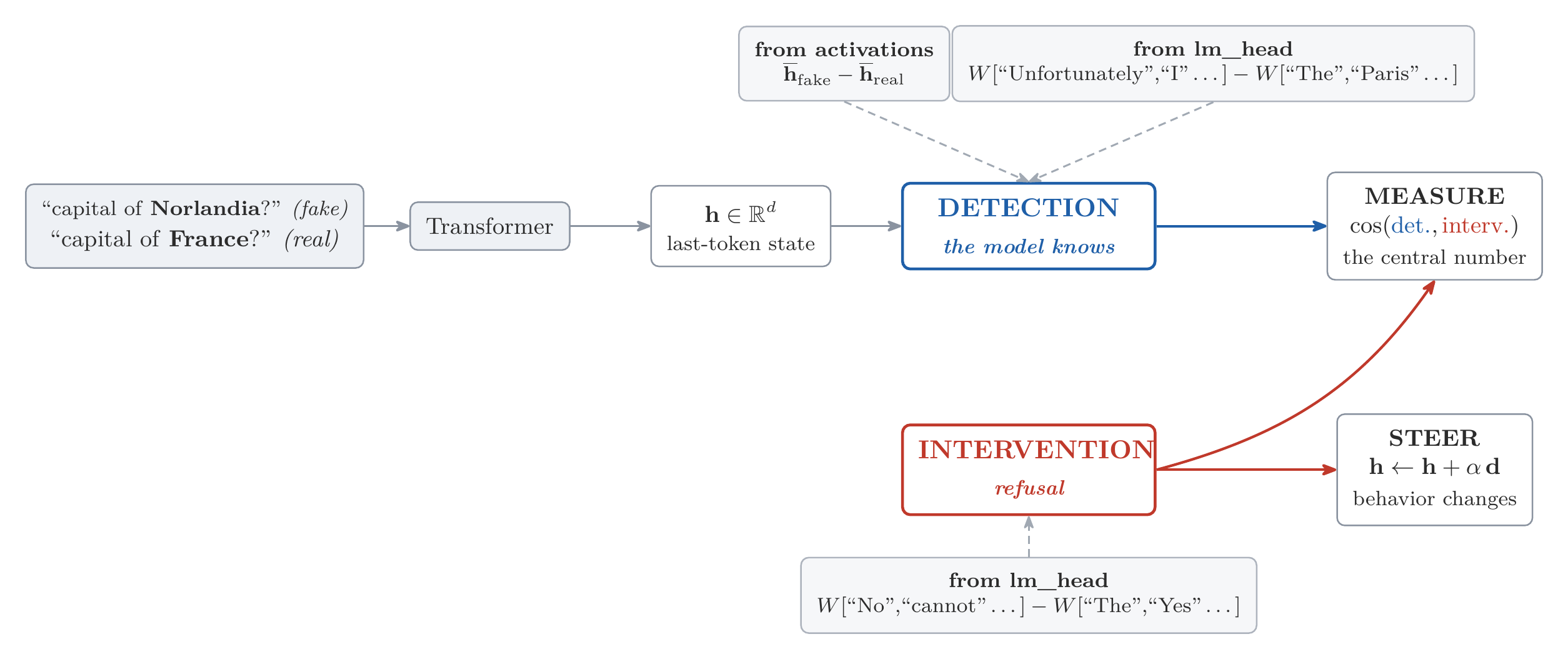}
\caption{\textbf{The method in one view.} From the last-token residual
state \(\mathbf{h}\) we build two directions: a \emph{detection}
direction --- whether the model internally registers an entity as fake,
obtainable either from activations (difference-in-means) or from the
output vocabulary (\protect\texttt{lm\_head}) --- and an
\emph{intervention} direction (refusal), read from
\protect\texttt{lm\_head} alone. We then \textbf{measure} the cosine
between them (the paper's central quantity) and \textbf{steer} by adding
a direction to \(\mathbf{h}\) during generation.}
\end{figure}

\subsubsection{2.1 Data-driven directions
(difference-in-means)}\label{data-driven-directions-difference-in-means}

To find the direction that distinguishes two behavioral conditions, we
collect residual-stream activations \(\mathbf{h}\) at the last prompt
token for a set of inputs from each condition, then subtract the means.
This difference-in-means (DiM) approach --- used by Arditi et al.~(2024)
for refusal and by Zou et al.~(2023) for representation engineering ---
requires no human judgment and captures whatever linear signal best
separates the two conditions.

\[\mathbf{d}_{\text{DiM}} = \text{normalize}\left(\text{mean}(\mathbf{h}_{\text{condition A}}) - \text{mean}(\mathbf{h}_{\text{condition B}})\right)\]

We evaluate detection quality with a logistic regression probe trained
on the same activations: an AUC of 1.000 means the two conditions are
perfectly linearly separable at that layer --- the model's internal
state already encodes the distinction without any ambiguity.

\subsubsection{2.2 Hand-picked directions (from the output
vocabulary)}\label{hand-picked-directions-from-the-output-vocabulary}

Rather than deriving a direction from activations, we read it from the
model's output weights (Turner et al., 2023; Zou et al., 2023). The
unembedding matrix \(W_{\text{lm\_head}}\) maps each residual-stream
vector to token logits, so the row for a token \emph{is} the
residual-space direction that promotes that token. A single token's row,
though, mostly points along what is common to all tokens; to isolate a
\emph{behavioral} axis we take a \textbf{contrast} --- the average row
of tokens typical of one response, minus the average row of tokens
typical of the opposite response. This is why each hand-picked direction
needs \textbf{two} token sets, a promoted set \(\mathcal{A}\) and a
contrasted set \(\mathcal{B}\):

\[\mathbf{d}_{\text{HP}} = \text{normalize}\left(\text{mean}(W_{\text{lm\_head}}[t_i \in \mathcal{A}]) - \text{mean}(W_{\text{lm\_head}}[t_j \in \mathcal{B}])\right)\]

The intuition: the \textbf{first token} the model emits already betrays
where a response is heading --- a cautious opener like
\emph{``Unfortunately''} precedes a non-answer, a confident
\emph{``The''} precedes a committed claim. We build two such directions.
The \textbf{detection direction} (hand-picked) promotes cautious openers
\(\mathcal{U}\) = \{``I'', ``Unfortunately'', ``There'', ``It'',
``This''\} and contrasts confident or fabricating ones \(\mathcal{C}\) =
\{``The'', ``In'', ``Paris'', ``Tokyo'', ``1''\}; the set
\(\mathcal{C}\) is not arbitrary --- ``The'' and ``In'' are the most
frequent first tokens when the model fabricates (``The capital of
Norlandia is\ldots{}''), and ``Paris'', ``Tokyo'' are entity names it
states outright, all read off its own output distribution. The
\textbf{intervention direction} --- because it promotes refusal tokens,
this is the \emph{refusal direction} of Arditi et al.~(2024) ---
promotes refusal openers \(\mathcal{R}\) = \{``No'', ``cannot'',
``doesn't'', ``I''\} and contrasts compliant ones \(\mathcal{O}\) =
\{``The'', ``Yes'', ``is'', ``It''\}.

\subsubsection{2.3 Detecting the fake/real
distinction}\label{detecting-the-fakereal-distinction}

To gauge how readily the model separates fake from real entities, we
read out the distinction six independent ways, spanning supervised
probes and label-free readouts (results in Section 4.1):

\begin{itemize}
\tightlist
\item
  \textbf{Linear probe} (supervised): logistic regression on the
  residual stream at each layer, leave-one-out cross-validated and
  reported as AUC, following standard probing methodology (Belinkov,
  2022).
\item
  \textbf{Single attention head} (supervised): the output of one head
  (L9 H2) scored for fake vs real.
\item
  \textbf{Logit lens --- top-5 entropy} (unsupervised): we project the
  residual stream at each layer through the unembedding matrix to read
  the model's evolving next-token prediction; the entropy of its top-5
  tokens measures predictive uncertainty.
\item
  \textbf{Single MLP neuron} (unsupervised): the activation of one
  neuron (N578, L15), read directly.
\item
  \textbf{SAE feature} (unsupervised): one feature (F15356) of the Gemma
  Scope sparse autoencoder (Google DeepMind, 2024) --- a dictionary that
  decomposes activations into monosemantic features --- that fires
  selectively on unrecognized entity names.
\item
  \textbf{Embedding norm} (unsupervised): the L2 norm of the entity
  token's input embedding, lower for unknown entities --- a label-free
  signal available before any computation.
\end{itemize}

All six are passive measurements: they show which layers carry the
distinction, not whether that signal causally drives behavior.

\subsubsection{2.4 Activation transplant}\label{activation-transplant}

We capture the full residual stream from a ``source'' forward pass (a
prompt that produces the desired behavior) and inject it into a
``target'' forward pass at specific layers. If the target adopts the
source's behavior while preserving its own content, the behavior is an
\emph{overlay} --- a separable signal that rides on top of the content
computation and can be transplanted cleanly.

\subsubsection{2.5 Causal intervention
protocol}\label{causal-intervention-protocol}

To test whether a candidate direction \(\mathbf{d}\) causally governs a
behavior, we add \(\alpha \cdot \mathbf{d}\) to the residual stream at
target layers during generation and measure behavioral change. The
scalar \(\alpha\) controls the intervention strength. To rule out
confounds, we run two controls: 100 random directions at matched norm,
and 10 semantically irrelevant token-pair directions. A direction is
considered causally relevant only if its effect significantly exceeds
both control baselines.

\subsubsection{2.6 Signal decomposition}\label{signal-decomposition}

At each layer, we decompose the residual stream update into attention
and MLP contributions using hooks at pre-feedforward and
post-feedforward positions, then project each contribution onto a
candidate direction to measure how much each sublayer carries the
behavioral signal. Gemma 2's four layernorms per layer allow clean
decomposition (Appendix A.5).

\subsubsection{2.7 Stimuli}\label{stimuli}

\textbf{General set (N=100).} The main stimulus set consists of 50
fake-entity questions across four thematic categories (capitals,
science, culture, geography) and 50 matched real-entity controls. All
fake entities are clearly non-existent by name (e.g., \emph{``What is
the capital of Norlandia?''}). This set is used for detection
experiments (Section 4.1), the geometric bottleneck analysis (Section
4.3), and signal decomposition (Section 5). The full list is in Appendix
A.3.

\textbf{Stress test (N=115).} Section 6.3 uses a harder, separate set
designed to probe the limits of the ROT-15° intervention (the rotated
intervention direction introduced in Section 6). We fixed these
categories from each stimulus's structural properties, before any
intervention experiment. To test whether the intervention generalizes
beyond easy cases, we split the stimuli by difficulty:

\begin{itemize}
\tightlist
\item
  \textbf{Type 1 fake} (N=30): entities with obviously invented names,
  where the phonology alone signals non-existence (e.g., \emph{``What is
  the capital of Karvistan?''}).
\item
  \textbf{Type 2 fake} (N=30): fabricated facts phrased around
  plausible-sounding dates, numbers, or institutions (e.g., \emph{``Who
  signed the Treaty of Valmora?''}). No phonological signal marks these
  as fake --- only world knowledge would reveal them.
\item
  \textbf{Real easy} (N=25): well-known real-world facts with
  unambiguous answers (e.g., \emph{``What is the capital of France?''}).
\item
  \textbf{Real obscure} (N=20): genuine but uncommon facts the model may
  not know confidently (e.g., \emph{``What is the capital of Nauru?''}),
  used to test whether the intervention causes over-refusal on hard but
  real questions.
\item
  \textbf{Real tricky-sounding} (N=10): real entities whose names
  superficially resemble invented ones (e.g., \emph{``What is the
  capital of Vanuatu?''}), the primary false-positive stress test.
\end{itemize}

\textbf{Format set.} Format experiments use 32 arithmetic queries with
matched fencing/no-fencing prompts (e.g., \emph{``What is 73+28?''}),
intentionally trivial to isolate the behavioral signal from content
complexity.

\begin{center}\rule{0.5\linewidth}{0.5pt}\end{center}

\subsection{3. Format: Where Detection Equals
Control}\label{format-where-detection-equals-control}

We first establish the positive case --- a behavior where detection and
intervention are geometrically aligned.

When asked to produce JSON, Gemma 2-2B-it wraps output in markdown code
fences 100\% of the time. Adding an explicit anti-fencing prompt
produces clean JSON 100\% of the time.

\subsubsection{3.1 The format direction}\label{the-format-direction}

Logit lens analysis --- projecting the residual stream at each layer
through the unembedding matrix to read the model's evolving token
prediction (Section 2.3) --- reveals the decision between \texttt{\{}
and \texttt{\textasciigrave{}\textasciigrave{}\textasciigrave{}} occurs
at layers 20--25, with peak divergence at layer 24. The format
direction:

\[\mathbf{d}_{\text{format}} = \text{normalize}(W_{\text{lm\_head}}[\text{`\{'}] - W_{\text{lm\_head}}[\text{`\textasciigrave\textasciigrave\textasciigrave'}])\]

\subsubsection{3.2 One axis both detects and
controls}\label{one-axis-both-detects-and-controls}

For format, detection and intervention are the \textbf{same direction}
--- not two directions that happen to align, but one vector doing both
jobs. The \texttt{lm\_head} format direction (Section 3.1) detects:
projecting the last-token residual onto it separates fencing from clean
output. And it controls: adding \(\mathbf{d}_{\text{format}}\) at
L20--25 with \(\alpha = 3.5\) eliminates fencing on 100\% of queries ---
a sharp sigmoid (100\% fencing below \(\alpha = 2.5\), 0\% above
\(\alpha = 3.5\)), 100\% valid JSON, 100\% correct answers, at an
intervention magnitude of 0.6\% of the activation norm. The direction
that \emph{separates} fencing from clean output is, literally, the one
that \emph{removes} it.

A caveat on what this does \emph{not} show: a \emph{data-driven}
direction built from fencing vs non-fencing activations is a different
vector, near-orthogonal to this one --- the two constructions do not
coincide. What makes format the aligned case is that a single
output-vocabulary direction serves both roles, not that independent
constructions agree. (This matters later: for hallucination no single
direction does both, and Section 8 shows why a cosine between
separately-built directions is not, by itself, a steerability test.)

Controls: 0/100 random directions and 0/10 irrelevant token-pair
directions affect fencing.

\subsubsection{3.3 The transplant test}\label{the-transplant-test}

Activations from a clean-JSON forward pass injected into a fencing
forward pass at L20--25 eliminate fencing 10/10 times across Gemma 2-2B
and Gemma 3-1B. Format is a \textbf{substitutable activation pattern}
--- the same content is rendered in either format by swapping a
localized state.

\subsubsection{3.4 Key observation}\label{key-observation}

Format \textbf{behaves as an overlay}: an output-rendering choice
imposed on top of content, separable enough to be swapped wholesale
(Section 3.3) and carried by a single axis that both detects and
controls it (Section 3.2). This is the aligned baseline: detection and
intervention are not two close directions but one and the same. Section
4 turns to hallucination, where they are two genuinely different
directions and the same setup gives a very different answer.

\begin{center}\rule{0.5\linewidth}{0.5pt}\end{center}

\subsection{4. Hallucination: Where Detection and Control
Diverge}\label{hallucination-where-detection-and-control-diverge}

\subsubsection{4.1 The model detects its own uncertainty by every
method}\label{the-model-detects-its-own-uncertainty-by-every-method}

The model separates fake from real entities by every detector we apply
(Section 2.3). All succeed:

\newpage

{\def\LTcaptype{none} 
\begin{longtable}[]{@{}llll@{}}
\toprule\noalign{}
Detector & Type & Where & Performance \\
\midrule\noalign{}
\endhead
\bottomrule\noalign{}
\endlastfoot
Logit lens (top-5 entropy) & Unsupervised & Layer 25 & AUC = 0.913 \\
Linear probe & Supervised & Layer 5+ & AUC = 1.000 \\
Single MLP neuron (N578) & Unsupervised & Layer 15 & Accuracy = 88\% \\
SAE feature (F15356) & Unsupervised & Layer 22 & 88\% fake / 0\% real \\
Embedding norm & Unsupervised & Input & Accuracy = 83\% \\
Single attention head (L9 H2) & Supervised & Layer 9 & AUC = 1.000 \\
\end{longtable}
}

The signal is \textbf{graded, not binary}: beyond telling fake from
real, the model separates \emph{obscure} real entities from
\emph{common} ones (L9 H2, AUC = 0.993), so its detector tracks how well
it knows an entity, not merely whether the entity exists. We are not the
first to see this. Kadavath et al.~(2022) found that language models
broadly represent calibrated uncertainty about their own knowledge, and
Azaria \& Mitchell (2023) showed that internal activations classify
model-generated falsehoods with near-perfect accuracy. Our contribution
relative to both is the systematic measurement of the \emph{causal} gap:
perfect detection coexisting with failed intervention. One might worry
that a probe in 2304 dimensions separates on prompt-surface features
rather than epistemic state. That it also separates two classes of
\emph{real} entities --- common from obscure, surface-matched and
differing only in how well the model knows them --- argues against a
purely superficial cut.

\subsubsection{4.2 Intervention: detection directions are weak or
backfire}\label{intervention-detection-directions-are-weak-or-backfire}

\textbf{Data-driven (DiM):} Difference-in-means achieves AUC = 1.000 for
detection --- the model's own activations separate fake from real
perfectly, with no human input. But DiM goes in the \textbf{wrong
direction} for intervention: applying it \emph{reduces} refusals,
increasing fabrication. DiM top tokens are semantically mixed, capturing
prompt-structure differences rather than the behavioral mechanism.

\textbf{Hand-picked (HP):} The hand-picked detection direction (Section
2.2), used as an intervention, \textbf{only partially works} --- at
\(\alpha = 15\) on L20--25, fabrication on the 100 fake entities drops
from \textasciitilde70\% to \textasciitilde40\% (\(p < 0.001\)), with 0
false refusals on the 50 real questions (the set is N=150: 50 Type 1 +
50 Type 2 fake + 50 real). That even the detection direction, pushed
directly, recovers only part of the behavior is exactly what the
near-orthogonality predicts; the intervention direction has stronger
causal power.

\textbf{Other methods that fail:} - SAE feature amplification (F15356,
up to 50×): no change in refusal (small collateral --- 2 real answers
lost). - Neuron ablation (top 500 discriminative neurons at L15): no
behavioral change.

The core pattern: detection succeeds by every method tested; only
output-vocabulary directions have causal power, and even they are
limited.

\subsubsection{4.3 The geometric
bottleneck}\label{the-geometric-bottleneck}

We measure the cosine between a detection direction and the intervention
direction. Detection, though, is not a single direction but an
\textbf{equivalence class}: the hand-picked and the data-driven (DiM)
detection directions both separate fake from real at AUC \(\approx\) 1,
yet are nearly orthogonal to \emph{each other} (\(\cos = 0.11\), about
83°) --- a whole family of directions detects, not one. We therefore
measure the gap to the intervention direction for \emph{both} of them:

\[\cos(\mathbf{d}_{\text{det}}, \mathbf{d}_{\text{ref}}) = 0.12, \qquad \cos(\mathbf{d}_{\text{DiM}}, \mathbf{d}_{\text{ref}}) = -0.06\]

Both are far from the \(\cos \approx 1\) that alignment would require.
The two values are not, however, the same measurement, and it is worth
being precise about their scale: in 2304 dimensions two unrelated
directions sit at \(\cos \approx 1/\sqrt{2304} = 0.02\) by chance. The
HP--HP value of 0.12 is a \emph{small but reproducible positive} ---
about 6× that floor, and consistent across models (§7); the cross-method
−0.06 is essentially at the floor. So neither approaches alignment, but
only the first is a real (if weak) signal --- and §8 shows why neither,
on its own, serves as a steerability test. This is the central result:
\textbf{every detection direction we construct is far from aligned with
the intervention direction.} The first cosine compares two directions
read from lm\_head; the second compares a direction derived purely from
\emph{activations} --- no lm\_head, no hand-chosen tokens --- against
the intervention direction. That an lm\_head-free construction yields
the same near-zero result rules out the most natural concern: that the
gap is an artifact of lm\_head geometry or token selection. Two further
checks corroborate it:

\begin{enumerate}
\def\labelenumi{\arabic{enumi}.}
\item
  \textbf{Cross-model convergence.} Three different lm\_head matrices
  with three different tokenizers (SentencePiece, BPE×2) all yield
  \(\cos \in [0.12, 0.20]\) (Section 7.1). If the measurement were an
  artifact of token selection, it would not converge across unrelated
  vocabularies.
\item
  \textbf{Base model replication.} \(\cos = 0.1197\) (base) vs
  \(0.1200\) (instruction-tuned) --- the same geometry in the same
  lm\_head, before and after alignment (Section 7.5).
\end{enumerate}

\textbf{Why the detection signal doesn't produce refusal:}

The quantities below are \emph{projections} of the last-prompt-token
residual stream at L25 onto a unit-norm direction --- the signed
component of the activation along that direction, in the model's native
activation units. We call the difference between the mean projection for
fake vs real entities the \emph{gap}.

The detection gap between fake and real entities at L25 is +49.8 --- an
enormous signal, 6.9× the mean gap of 2000 random unit directions. But
the refusal gap along the same axis is only −24.9, in the \textbf{wrong
direction}.

\begin{enumerate}
\def\labelenumi{\arabic{enumi}.}
\item
  \textbf{The entity-copy mechanism dominates.} Logit lens at L25: for
  \emph{``Capital of Norlandia?''} the model predicts the entity
  fragment `Nor' with \textasciitilde99\% probability; for real
  questions it predicts `The' with comparable confidence. The detection
  component is a small slice of the residual: the part orthogonal to the
  detection direction --- dominated by the entity copy, though not
  \emph{only} entity copy --- is roughly an order of magnitude larger
  (median \textasciitilde12× across the fake set). This domination is
  \textbf{not specific to fakes} --- it holds, even more strongly, for
  real entities, so it is a general property of the output mapping
  rather than a signature of the failure case; the detection direction
  is simply a thin slice of what drives the output. The amplification of
  the salient entity token is related to the copy-suppression motif
  characterized in GPT-2 by McDougall et al.~(2023); here the mechanism
  operates in the \emph{opposite} direction, amplifying rather than
  suppressing entity copies at late layers.
\item
  \textbf{The detection direction actively harms refusal tokens.}
  Decomposing logit contributions on fake entities, the detection
  direction pushes \emph{down} on refusal tokens (e.g.~``fictional'')
  and \emph{up} on fabrication openers (``The''). The direction that
  \emph{perfectly detects} the fake entity pushes the output
  \emph{toward} fabrication.
\item
  \textbf{Removing the detection component doesn't change predictions.}
  With the detection projection removed from the residual stream,
  ``Norlandia'' still maps to `Nor' with essentially unchanged
  probability. The detection signal is invisible to the output mapping.
\end{enumerate}

\begin{center}\rule{0.5\linewidth}{0.5pt}\end{center}

\subsection{5. Signal Autopsy: Where the Detection Signal
Lives}\label{signal-autopsy-where-the-detection-signal-lives}

\subsubsection{5.1 MLP dominates
attention}\label{mlp-dominates-attention}

We decompose the residual stream layer-by-layer into attention and MLP
contributions, projecting each onto the detection direction.

\textbf{Cumulative gap} (fake − real projection on detection direction):
- MLP: +42.3 - Attention: +7.5 - \textbf{Ratio: MLP carries 5.66× more
signal}

The MLP contribution is spread across the late layers (L18--L25 each add
roughly +5 to +7 to the gap); no single layer dominates. Attention
mildly opposes the signal at L17 (−0.92) and L21 (−0.76), but never
strongly. This MLP dominance is consistent with Geva et al.~(2023), who
showed that factual association retrieval in autoregressive models is
driven primarily by mid-layer MLP computations; our finding extends this
to the encoding of epistemic uncertainty about entity existence.

\subsubsection{5.2 Attention at L17+ is
dispensable}\label{attention-at-l17-is-dispensable}

We perform franken-forward experiments, zeroing attention output at
specific layers:

{\def\LTcaptype{none} 
\begin{longtable}[]{@{}
  >{\raggedright\arraybackslash}p{(\linewidth - 4\tabcolsep) * \real{0.1618}}
  >{\raggedright\arraybackslash}p{(\linewidth - 4\tabcolsep) * \real{0.3971}}
  >{\raggedright\arraybackslash}p{(\linewidth - 4\tabcolsep) * \real{0.4412}}@{}}
\toprule\noalign{}
\begin{minipage}[b]{\linewidth}\raggedright
Condition
\end{minipage} & \begin{minipage}[b]{\linewidth}\raggedright
Fake refuse (baseline 60\%)
\end{minipage} & \begin{minipage}[b]{\linewidth}\raggedright
Real correct (baseline 100\%)
\end{minipage} \\
\midrule\noalign{}
\endhead
\bottomrule\noalign{}
\endlastfoot
Kill L17+L21 (fighters) & 60\% & 100\% \\
Kill L20-25 (late) & 65\% & 100\% \\
MLP-only L13-25 & 5\% (destroyed) & 0\% (destroyed) \\
\textbf{MLP-only L17-25} & \textbf{75\%} & \textbf{95\%} \\
\end{longtable}
}

Attention through L16 is essential for text coherence. From L17 onward,
it is dispensable --- removing it actually \emph{improves} honesty by
+15 percentage points with minimal degradation of real answers.

\subsubsection{5.3 Per-layer signal is too small for
gating}\label{per-layer-signal-is-too-small-for-gating}

We attempted to build a self-intervention gate: read the MLP output at
one layer, project onto the detection direction, use the value to
modulate attention at later layers. 9 conditions tested, all equal to
baseline.

\textbf{Root cause}: the detection gap at any single MLP layer output is
\textasciitilde0.01 --- far too small to use as a gate. The cumulative
gap of \textasciitilde50 exists only as the \emph{sum} across all
layers. No single layer is a reliable thermometer.

\begin{center}\rule{0.5\linewidth}{0.5pt}\end{center}

\subsection{6. Bridging the Gap: The 15°
Rotation}\label{bridging-the-gap-the-15-rotation}

\subsubsection{6.1 Method}\label{method}

Since the detection direction (\(\mathbf{d}_{\text{det}}\)) and the
intervention direction (\(\mathbf{d}_{\text{ref}}\)) are nearly
orthogonal (\(\cos = 0.12\)), we can rotate from the detection direction
toward the intervention direction using Gram-Schmidt orthogonalization:

\[\mathbf{d}_{\text{rot}}(\theta) = \cos\theta \cdot \mathbf{d}_{\text{det}} + \sin\theta \cdot \mathbf{d}_{\text{ref}\perp}\]

where \(\mathbf{d}_{\text{ref}\perp}\) is the component of the
intervention direction orthogonal to the detection direction.

\subsubsection{6.2 Exploratory sweep}\label{exploratory-sweep}

We sweep \(\theta\) from 0° to 90° on a curated subset of 20 fake + 20
real entities. \textbf{Every row below has \(\alpha = 15\) applied} ---
the sweep compares rotation angles against each other, not intervention
vs no-intervention. The no-intervention baseline on this curated subset
is 60\% refusal (12/20), \emph{not} the 85\% in the 0° row (which
already has \(\alpha = 15\)):

{\def\LTcaptype{none} 
\begin{longtable}[]{@{}lll@{}}
\toprule\noalign{}
\(\theta\) & Fake refuse (N=20) & Real correct (N=20) \\
\midrule\noalign{}
\endhead
\bottomrule\noalign{}
\endlastfoot
0° (pure detection) & 85\% & 100\% \\
\textbf{15°} & \textbf{90\%} & \textbf{100\%} \\
30° & 85\% & 95\% \\
45° & 90\% & 95\% \\
60° & 85\% & 100\% \\
90° (pure refusal\(\perp\)) & 80\% & 100\% \\
\end{longtable}
}

At N=20, most cell differences are within sampling noise (1--2 samples).
This sweep is underpowered to distinguish specific angles --- it serves
to show that the range is broadly effective and to select 15° as a
candidate for validation, balancing 97\% detection sensitivity
(\(\cos 15° = 0.966\)) with 26\% action component
(\(\sin 15° = 0.259\)).

\subsubsection{6.3 Validation on held-out hard
cases}\label{validation-on-held-out-hard-cases}

We validate ROT-15° at \(\alpha = 15\) on the stress test (N=115: 30
Type 1 fake, 30 Type 2 fake, 55 real split into
easy/obscure/tricky-sounding; see §2.7 for category definitions),
distinct from the sweep set.

{\def\LTcaptype{none} 
\begin{longtable}[]{@{}lll@{}}
\toprule\noalign{}
Group & Baseline (no intervention) & ROT-15° \(\alpha=15\) \\
\midrule\noalign{}
\endhead
\bottomrule\noalign{}
\endlastfoot
Type 1 (obvious fake) & 40\% refuse & \textbf{73\% refuse} \\
Type 2 (subtle fake --- dates, numbers) & 13\% refuse & \textbf{60\%
refuse} \\
Real easy (N=25) & 100\% correct & 100\% correct \\
Real obscure (N=20) & 100\% correct & 95\% correct \\
Real tricky-sounding (N=10) & 60\% correct & 60\% correct \\
\textbf{False positives} & 0/55 & \textbf{1/55 (1.8\%)} \\
\end{longtable}
}

The baselines differ from the sweep (Section 6.2) because this is a
harder stimulus set: Type 1 entities include cases that the model
already partially refuses (40\% baseline), while Type 2 entities
(fabricated dates, numbers) are nearly always accepted (13\% baseline).
The sweep used curated ``easy'' fakes where even the pure detection
direction achieves 85\%.

Type 2 hallucinations --- where the model invents dates, numbers, and
people with full confidence --- go from 13\% to 60\% refusal. The single
false positive (Bosch/Garden of Earthly Delights) is an arguably
borderline case.

At \(\alpha = 10\): zero false positives, Type 1 40\%→57\%, Type 2
13\%→33\%.

\subsubsection{6.4 Interpretation}\label{interpretation}

A 15° rotation in 2304-dimensional space partially bridges the gap
between detecting and acting on hallucination: it recovers much of the
refusal behavior (Type 2: 13\%→60\%) but not all, and the sweep (Section
6.2) is underpowered to pin the angle down precisely. Still, the model
has all the information needed to refuse --- the bottleneck is that this
information lives along a direction nearly orthogonal to the
intervention direction.

\begin{center}\rule{0.5\linewidth}{0.5pt}\end{center}

\subsection{7. Cross-Model Replication}\label{cross-model-replication}

\subsubsection{7.1 Four-model
triangulation}\label{four-model-triangulation}

We replicate the lm\_head bottleneck analysis --- the same
\(\cos(\mathbf{d}_{\text{det}}, \mathbf{d}_{\text{ref}})\) and
gap-vs-random measurement of Sections 2 and 4.3, with both directions
built from each model's own lm\_head --- on Llama-3.2-1B-Instruct,
Qwen-2.5-1.5B-Instruct, and Gemma 2-9B-it.

{\def\LTcaptype{none} 
\begin{longtable}[]{@{}llllll@{}}
\toprule\noalign{}
Model & Layers & dim & cos & Signal ratio & Behavior \\
\midrule\noalign{}
\endhead
\bottomrule\noalign{}
\endlastfoot
Gemma 2-2B-it & 26 & 2304 & \textbf{0.12} & 6.9× & Fabricates
(\textasciitilde70\%) \\
Llama 3.2-1B & 16 & 2048 & \textbf{0.20} & 12.5× & Refuses (100\%) \\
Qwen 2.5-1.5B & 28 & 1536 & \textbf{0.16} & 2.7× & Mixed (40\%
refuse) \\
Gemma 2-9B-it & 42 & 3584 & \textbf{0.13} & 12.8× & Fabricates \\
\end{longtable}
}

\emph{cos} = \(\cos(\mathbf{d}_{\text{det}}, \mathbf{d}_{\text{ref}})\),
the cosine between detection and intervention directions. \emph{Signal
ratio} = gap between fake and real projections on the detection
direction, relative to a random baseline.

All four show near-orthogonality (\(\cos \in [0.12, 0.20]\)),
corresponding to angles of 78°--83° between the detection and
intervention directions. These are small values --- but consistently
\emph{above} the chance floor (\textasciitilde0.02 in these dimensions,
§4.3) and reproducible across unrelated tokenizers; it is this
\textbf{consistency of a small positive}, not its magnitude, that is the
invariant. Three different model families, three different tokenizers
(SentencePiece, BPE×2), and two scales within the same family (2B vs 9B)
confirm the pattern. The notably smaller gap/random ratio for Qwen (2.7×
vs \textasciitilde7--13× for Gemma and Llama) reflects Qwen's larger
residual norms absorbing the perturbation rather than a weaker detection
signal --- consistent with Qwen's no-collapse behavior under large
\(\alpha\) (Section 7.4).

\subsubsection{7.2 Token alignment is
universal}\label{token-alignment-is-universal}

Ranking the whole vocabulary by its alignment with the (fixed) detection
direction, the top tokens are the same across all three models despite
completely different tokenizers --- and they are not the tokens the
direction was built from:

\begin{itemize}
\tightlist
\item
  \textbf{Favored} (uncertainty): ``Unfortunately'', ``There'',
  ``Sadly'', ``It'', ``I'', ``Sorry''
\item
  \textbf{Disfavored} (confidence): ``Paris'', ``Tokyo'', numbers, city
  names
\end{itemize}

Qwen additionally includes the Chinese-script equivalents of Paris and
Tokyo among its disfavored tokens --- the same semantic structure
expressed through a different script and tokenizer. The direction
encodes a language-modeling-level concept, not a tokenizer artifact.

\subsubsection{7.3 Causal power is
bidirectional}\label{causal-power-is-bidirectional}

On Gemma 2-2B-it, ROT-15° reaches 73\% and 60\% refusal on the held-out
stress test (Section 6.3). Does the same direction --- built from each
model's own lm\_head --- also have causal power on other families? On
small exploratory sets (15 fake + 15 real per model), it does:

{\def\LTcaptype{none} 
\begin{longtable}[]{@{}lllll@{}}
\toprule\noalign{}
Model & Baseline fake refuse & Best \(\alpha\) & Effect & Real damage \\
\midrule\noalign{}
\endhead
\bottomrule\noalign{}
\endlastfoot
Llama 3.2-1B & 100\% & \(\alpha = -1\) & 80\% \textbf{fabricate} &
0/15 \\
Qwen 2.5-1.5B & 40\% & \(\alpha = +5\) & 93\% refuse & 0/15 FP \\
\end{longtable}
}

In each case \(\alpha\) is pushed toward the behavior the model does
\emph{not} default to: positive to induce refusal (Qwen, as for Gemma in
Section 6.3), negative to induce fabrication (Llama). Llama already
refuses everything, so the negative \(\alpha\) selectively induces
fabrication on fake entities while preserving real answers --- the same
direction steers behavior both ways, only from a different baseline.

Llama diverges from the other models. Despite the same near-orthogonal
geometry (\(\cos = 0.20\)), it refuses fake entities by default ---
detection and action agree in outcome here, where on Gemma they come
apart. We examine what this means for the central claim in the
Discussion (Section 8).

\(\alpha\) magnitude does not transfer: Gemma needs \(\alpha = 15\),
Llama needs \(|\alpha| = 1\), Qwen needs \(\alpha = 5\). The direction
generalizes; the scale must be recalibrated per model.

\subsubsection{7.4 Collapse signatures are
interpretable}\label{collapse-signatures-are-interpretable}

At excessive \(\alpha\): - Gemma: ``Paris Paris Paris\ldots{}''
(confidence token loop) - Llama: ``TokTokTok\ldots{}'' --- ``Tok'' is
the BPE prefix for ``Tokyo'' (entity-copy mechanism takes over) - Qwen:
no collapse across entire tested range (large residual norms absorb the
perturbation)

The Llama collapse reveals the entity-copy mechanism (Section 4.3) in a
different model: when the detection signal is ablated past the sweet
spot, the entity-copy component becomes dominant, producing an
interpretable behavioral collapse rather than random noise.

\subsubsection{7.5 Instruction tuning leaves the lm\_head geometry
unchanged}\label{instruction-tuning-leaves-the-lm_head-geometry-unchanged}

We measure \(\cos(\mathbf{d}_{\text{det}}, \mathbf{d}_{\text{ref}})\) on
the Gemma 2-2B base model (no instruction tuning):

{\def\LTcaptype{none} 
\begin{longtable}[]{@{}ll@{}}
\toprule\noalign{}
Model & \(\cos(\mathbf{d}_{\text{det}}, \mathbf{d}_{\text{ref}})\) \\
\midrule\noalign{}
\endhead
\bottomrule\noalign{}
\endlastfoot
Gemma 2-2B (base) & \textbf{0.1197} \\
Gemma 2-2B-it & \textbf{0.1200} \\
\end{longtable}
}

Difference: 0.0003. The instruction tuning does not touch the lm\_head
geometry.

What alignment \emph{does} change is where the activations land along
the (fixed) detection direction. In the base model the ordering is
backwards --- real entities project further along it, as if \emph{more}
uncertain than the fakes (gap −23.2). Instruction tuning flips it the
sensible way: now the fake entities are the ones that look uncertain
(gap +49.8). The geometry is the same in both models; tuning leaves the
axis untouched and moves the activations along it.

The detection direction is multilingual already in pretraining: top
favored tokens include ``Unfortunately'' (English, 0.809), ``purtroppo''
(Italian, 0.460), ``malheureusement'' (French). The concept is baked in
by the language modeling objective.

\begin{center}\rule{0.5\linewidth}{0.5pt}\end{center}

\subsection{8. Discussion}\label{discussion}

\subsubsection{A consistent geometric
pattern}\label{a-consistent-geometric-pattern}

Our central finding is a precise geometric characterization of the
detection-intervention gap. Across three families, three tokenizers, and
two scales within the Gemma family (2B and 9B) --- with and without
instruction tuning --- the angle between the detection and intervention
directions consistently falls in the range of 78°--83°
(\(\cos \in [0.12, 0.20]\)), and is robust to whether the detection
direction is built from activations or from the output vocabulary.

This consistency is what we would expect if the gap reflects an
architectural regularity of autoregressive language modeling. The
pretraining objective optimizes lm\_head to predict the \emph{next
token}, not to map internal epistemic states to behaviorally appropriate
outputs. The model learns to represent uncertainty (detection) and
learns to produce tokens (action) as largely independent computations.
They share the same vector space but occupy nearly orthogonal subspaces
within it. Whether this pattern holds for encoder-decoder architectures,
models above 10B parameters, or non-factual behaviors remains an open
question.

\subsubsection{Detection and action are independent
faculties}\label{detection-and-action-are-independent-faculties}

The \(\cos\) stays near zero across all four models --- detection and
intervention sit on nearly orthogonal directions, invariantly --- yet
the behavioral consequence varies along a spectrum. Gemma detects that
an entity is fake (AUC = 1.000) and fabricates most of the time:
detection with little corresponding action. Qwen sits in the middle,
refusing about 40\% and fabricating the rest. Llama, with the identical
geometry (\(\cos = 0.20\)), refuses fakes outright. The same
near-orthogonal geometry coexists with outcomes from ``almost always
fabricate'' to ``always refuse'': the geometry does not determine the
behavior.

Llama sharpens this into a double dissociation. It cannot be routing its
refusal \emph{through} the detection direction --- that direction is as
orthogonal to refusal as Gemma's --- so its correct behavior must travel
a separate path. We thus have detection without action (Gemma) and
action that does not flow from detection (Llama), the standard pattern
for showing two functions are genuinely distinct. The model on which our
gap does not show up behaviorally is, in that sense, strong evidence
that detecting and acting are separate computations, not one faculty
that occasionally fails to connect.

How Llama achieves this is open. Its refusal is selective --- it answers
real entities and refuses fake ones --- so some fake/real signal does
reach the refusal output; yet the detection direction we measure is
orthogonal to that output. The routing must therefore run through a path
our linear \texttt{lm\_head} directions do not capture: a different
detection direction (detection is a class of directions, not one ---
Section 4.3), or a non-linear, multi-step route. Localizing it is left
to future work.

\subsubsection{The shortcut that fails: a weight-cosine is not a
steerability
oracle}\label{the-shortcut-that-fails-a-weight-cosine-is-not-a-steerability-oracle}

It is natural to hope this cosine doubles as an \emph{a priori}
steerability test --- high \(\cos\) (format) meaning the detection
direction is itself a control knob, low \(\cos\) (hallucination) meaning
one should steer a rotated direction instead. We tested that reading and
it does not hold, for three reasons that are themselves the result.

\textbf{First, alignment is the exceptional case, not a pole of a clean
scale.} Measured the same independent way --- a data-driven detector
against the intervention direction --- the cosine sits near the
high-dimensional chance level (\(\approx 1/\sqrt{2304} = 0.02\)) for
\emph{both} format and hallucination. Format's apparent
\(\cos \approx 1\) is not two constructions agreeing; it is one and the
same vector used in both roles (Section 3.2).

\textbf{Second, detection is not a single direction but a class.} The
hand-picked and data-driven detectors both reach AUC \(\approx 1\) yet
are themselves near-orthogonal (\(\cos = 0.11\); Section 4.3). ``The''
detection direction --- the single object a one-number diagnostic
presupposes --- does not exist; detection occupies a subspace.

\textbf{Third, what separates the steerable case from the unsteerable
one is functional, not geometric.} The test is whether the
\emph{intervention} direction --- the one that steers --- also works as
a \emph{detector}. For format it does: the same direction doubles as a
near-perfect detector (AUC \(\approx 1\)). For hallucination it does not
(AUC \(\approx 0.7\)) --- and note this is the \emph{intervention}
(refusal) direction; the dedicated \emph{detection} direction still
separates fake from real at AUC = 1.000 (§4.1). The model detects
perfectly, but the direction that \emph{acts} barely reads the behavior
it is supposed to act on. That difference --- whether the controlling
direction also reads the behavior --- is a fact about behavior under
intervention, invisible to a static angle.

What the cosine \emph{is} is a robust, weight-computable
\textbf{signature of the dissociation}, invariant across four models
(\(\cos \in [0.12, 0.20]\)) --- a description of how far detection sits
from control, not a predictor of how steerable a behavior is. The
rotation of Section 6 is best read in the same spirit: evidence that the
gap has geometric \emph{structure} (it can be partly bridged --- Type 2
refusal 13\%→60\%) rather than a turnkey fix, since it recovers only
part of the behavior.

\subsubsection{Alignment and residual-stream
geometry}\label{alignment-and-residual-stream-geometry}

If the lm\_head geometry is fixed by pretraining (\(\cos\) difference of
0.0003 between base and IT), how does alignment training produce models
that refuse unknown entities (Llama) vs fabricate (Gemma)? One
interpretation consistent with our data is that alignment works by
shifting \emph{residual-stream representations} --- moving where
fake-entity activations land relative to the fixed lm\_head geometry ---
rather than by modifying the geometry itself.

Testing this directly would require comparing activation trajectories
between base and instruction-tuned models across the full forward pass,
which we leave to future work. What we can say is that \(\alpha\) does
not transfer across models (Gemma needs 15, Llama needs \(|1|\), Qwen
needs 5), consistent with each model's residual stream starting at a
different point relative to the same geometric bottleneck. Scale matters
too: Gemma 2-9B-it reproduces the geometry (Section 7.1,
\(\cos = 0.13\)) but resists steering --- the same intervention is
suppressed rather than amplified at 9B --- so we report its geometry but
not a comparable intervention. Whether larger models systematically damp
this steering is left to future work.

\subsubsection{The entity-copy
bottleneck}\label{the-entity-copy-bottleneck}

In the output mapping, the residual orthogonal to detection ---
dominated by entity copy --- outweighs the detection component by
roughly an order of magnitude (median \textasciitilde12× on fakes, and
more on reals: the domination is general, not specific to the failure
case; Section 4.3), which explains why the detection signal, despite
being enormous (6.9× random), fails to influence behavior. The model's
output is dominated by a mechanism that copies the most salient entity
token from the context --- a pattern useful for factual recall but
catastrophic for unknown entities.

We document this dominance but do not explain its emergence from
pretraining. Why does entity copy outweigh the detection signal so
heavily in the output, and would the ratio change for non-entity-centric
questions? These are open questions for future work. The entity-copy
mechanism is visible in all three models (the ``TokTokTok'' collapse on
Llama), at strengths that vary by model.

\subsubsection{MLP as the signal
carrier}\label{mlp-as-the-signal-carrier}

The signal autopsy (Section 5) reveals an asymmetric division of labor:
MLP layers carry 5.66× more detection signal than attention, spread
across the late layers rather than concentrated in one. Attention at
L17+ can be entirely removed while \emph{improving} honesty. The
detection-action gap is not a coordination failure between attention and
MLP --- the signal is overwhelmingly in MLP, which feeds directly into
the residual stream, yet the output mapping still ignores it.

\subsubsection{Limitations}\label{limitations}

\textbf{Sample sizes.} Most experiments use 20--55 questions per
condition. Effects are large and consistent, but replication at larger
scale is warranted. The rotation sweep (Section 6.2, N=20) is
underpowered for fine-grained angle selection; we use it as an
exploratory step validated on the larger stress test (Section 6.3,
N=115).

\textbf{Model families.} Three families tested (Gemma, Llama, Qwen), all
decoder-only transformers. Coverage extends from 1B to 9B parameters via
Gemma 2-2B-it and Gemma 2-9B-it (\(\cos = 0.12\) and \(0.13\)
respectively). The consistent \(\cos \in [0.12, 0.20]\) across families
and scales is strong but not universal: replication on models above 10B
and on non-decoder architectures remains needed.

\textbf{Single behavior deep-dive.} The geometric analysis focuses on
hallucination about non-existent entities, with output format as a
positive control. Other behaviors may exhibit different angular
relationships; whether the angle varies systematically across a broader
range of behaviors --- from overlay to substrate --- is left to future
work.

\textbf{Entity-copy mechanism.} The order-of-magnitude dominance of
entity copy (median \textasciitilde12× on fakes, larger on reals) is
documented but unexplained, and is not specific to the failure case.
Whether it is specific to entity-centric questions or a general property
of the output mapping is untested. Explaining its emergence from
pretraining is an open question.

\textbf{lm\_head linearity.} Our direction construction relies on
lm\_head rows, which assumes approximate linearity of the unembedding.
This holds well at 1--3B parameters. Whether the orthogonality persists
when measured via nonlinear probes on larger models is an open question
--- the gap could be even more pronounced if the unembedding is less
faithful at scale.

\textbf{Stimulus difficulty tiers.} The Type 1/Type 2 and
easy/obscure/tricky-sounding categories in the stress test (Section 6.3)
are pre-specified and based on structural properties of the stimuli, not
on model behavior. However, the boundary between tiers is not always
sharp --- a human judge might reasonably classify some stimuli
differently. The reported per-tier results should be interpreted as
indicative of a difficulty gradient rather than as hard category
boundaries. The aggregate false-positive rate (1/55) is unaffected by
this concern.

\textbf{Non-identifiability.} Steering vectors have large equivalence
classes. We tested 5 direction variants per behavior and found the
separability contrast invariant across effective directions, but cannot
rule out that some untested direction breaks the pattern.

\begin{center}\rule{0.5\linewidth}{0.5pt}\end{center}

\subsection{9. Related Work}\label{related-work}

\textbf{Arditi et al.~(2024)} demonstrated that refusal is mediated by a
single direction on Llama 7B/13B --- an aligned case where detection
\(\approx\) control. \textbf{Kazemi et al.~(2026)} refined this to
single-neuron granularity: suppressing one MLP neuron suffices to bypass
safety alignment across 7 models from 1.7B to 70B (91.7\% attack success
on JailbreakBench, no training required). Crucially, the same neuron
serves as a near-perfect harmful-prompt detector (AUROC \(\approx\)
Llama-Guard-3-8B) --- an instance where detection \(\approx\)
intervention at the finest possible grain. Our finding that the angular
relationship is \textasciitilde83° for hallucination on Gemma 2-2B
provides the complementary negative case --- detection and control
occupying nearly orthogonal subspaces even where detection is perfect.

\textbf{Li et al.~(2023)} showed truth-telling directions from probing
can increase honesty. Our results suggest this works when the probed
direction also functions as the causal one (overlay) but fails when it
does not (substrate) --- though we find (Section 8) that a static cosine
does not by itself predict which case applies.

\textbf{Turner et al.~(2023)} formalized activation addition for
behavioral steering, building on \textbf{Subramani et al.~(2022)}, who
first extracted latent steering vectors from language models.
\textbf{Zou et al.~(2023)} proposed representation engineering as a
general framework. Our geometric analysis characterizes the two regimes
these approaches meet --- overlay (detection and control on a single
axis) and substrate (near-orthogonal) --- while showing (Section 8) that
the regime is not reliably read off a single weight-cosine.

\textbf{Marks \& Tegmark (2024)} characterized the geometry of truth
representations. Our work extends geometric analysis to the
\emph{relationship between} detection and intervention directions,
finding that high-quality truth detection does not guarantee steering
effectiveness.

\textbf{Park et al.~(2023)} and \textbf{Hernandez et al.~(2023)} provide
theoretical and empirical support for the linear representation
hypothesis --- that features are encoded as approximately linear
directions in the residual stream. Our direction construction (Section
2.2) relies on this assumption. The consistency of
\(\cos \in [0.12, 0.20]\) across three architectures with different
tokenizers provides further empirical support for the approximate
linearity of the unembedding.

\textbf{Burns et al.~(2023)} discovered latent knowledge via
unsupervised probing (CCS). Our finding that perfect probing accuracy
coexists with zero intervention effectiveness provides a mechanism for
\emph{why} latent knowledge remains latent: the routing from knowledge
to behavior fails when detection and action occupy orthogonal subspaces.

\textbf{Azaria \& Mitchell (2023)} showed that internal activations
classify model-generated false statements with near-perfect accuracy ---
an independent confirmation of the detection side of our finding. We add
the causal counterpart: that this internal signal does not propagate to
the output action.

\textbf{Rimsky et al.~(2024)} applied contrastive activation addition to
steer Llama 2 on safety-relevant behaviors, finding that single
directions effectively control outputs across a range of
alignment-relevant settings. Their results are consistent with our
format case (overlay: a single direction both detects and controls). Our
hallucination case (\(\cos \approx 0.12\)) identifies the failure mode
where single-direction steering is insufficient --- the detection
direction is nearly orthogonal to the intervention direction.

\textbf{Meng et al.~(2022)} located factual associations in GPT via
causal tracing. Our finding that factual knowledge is extremely robust
to steering (\(\alpha = 20\) needed to degrade real answers, vs
\(\alpha = 3.5\) for format) is consistent with their finding that facts
are stored in MLP weights.

\textbf{Heimersheim \& Nanda (2024)} recommended best practices for
activation patching, noting that path-patching and ablation can give
different results. Our own circuit-level ablations confirm this: heads
identified by path patching as carrying the ``known entity?'' signal
\emph{increase} fabrication when ablated --- detection \(\neq\)
intervention at the circuit level.

\textbf{Kaplan et al.~(2026)} showed that fine-tuning encourages
hallucinations through a forgetting mechanism. Our work addresses the
complementary case: even when the model has \emph{not} forgotten (AUC =
1.000), the routing from knowledge to behavior can fail. These are two
distinct failure modes.

\textbf{Yona et al.~(2026)} frame hallucinations as ``confident errors''
--- incorrect information delivered without appropriate uncertainty ---
and argue that \emph{faithful uncertainty} (aligning expressed
confidence with intrinsic knowledge) is the key path forward. Our
geometric analysis provides a mechanism for why this alignment fails:
the model's internal signal is accurate (AUC = 1.000) but nearly
orthogonal to the intervention direction (\(\cos \approx 0.12\)). The
gap is not epistemic but geometric.

\begin{center}\rule{0.5\linewidth}{0.5pt}\end{center}

\subsection{References}\label{references}

\begin{itemize}
\tightlist
\item
  Arditi, A., Obeso, O., Nanda, N., \& Mallen, J. (2024). Refusal in
  Language Models Is Mediated by a Single Direction. \emph{NeurIPS
  2024}. arXiv:2406.11717
\item
  Azaria, A. \& Mitchell, T. (2023). The Internal State of an LLM Knows
  When It's Lying. \emph{EMNLP 2023 Findings}. arXiv:2304.13734
\item
  Belinkov, Y. (2022). Probing Classifiers: Promises, Shortcomings, and
  Advances. \emph{Computational Linguistics}, 48(1). arXiv:2102.12452
\item
  Burns, C., Ye, H., Klein, D., \& Steinhardt, J. (2023). Discovering
  Latent Knowledge in Language Models Without Supervision. \emph{ICLR
  2023}. arXiv:2212.03827
\item
  Dubey, A. et al.~(2024). The Llama 3 Herd of Models. arXiv:2407.21783
\item
  Gemma Team (2024). Gemma 2: Improving Open Language Models at a
  Practical Size. arXiv:2408.00118
\item
  Geva, M., Bastings, J., Filippova, K., \& Globerson, A. (2023).
  Dissecting Recall of Factual Associations in Auto-Regressive Language
  Generation. \emph{EMNLP 2023}. arXiv:2304.14767
\item
  Google DeepMind (2024). Gemma Scope: Open Sparse Autoencoders
  Everywhere All At Once in Gemma 2. arXiv:2408.05147
\item
  Heimersheim, S. \& Nanda, N. (2024). Best Practices for Activation
  Patching. arXiv:2404.15255
\item
  Hernandez, E., Wattenberg, M., \& Andreas, J. (2023). Linearity of
  Relation Decoding in Transformer Language Models. \emph{ICLR 2024}.
  arXiv:2308.09124
\item
  Kadavath, S. et al.~(2022). Language Models (Mostly) Know What They
  Know. arXiv:2207.05221
\item
  Kaplan, G. et al.~(2026). Why Fine-Tuning Encourages Hallucinations
  and How to Fix It. arXiv:2604.15574
\item
  Kazemi, H., Chegini, A., \& Safi, M. (2026). A Single Neuron Is
  Sufficient to Bypass Safety Alignment in Large Language Models.
  arXiv:2605.08513
\item
  Li, K., Patel, O., Viégas, F., Pfister, H., \& Wattenberg, M. (2023).
  Inference-Time Intervention: Eliciting Truthful Answers from a
  Language Model. \emph{NeurIPS 2023}. arXiv:2306.03341
\item
  Marks, S. \& Tegmark, M. (2024). The Geometry of Truth: Emergent
  Linear Structure in Large Language Model Representations of True/False
  Datasets. \emph{COLM 2024}. arXiv:2310.06824
\item
  McDougall, C., Conmy, A., Rushing, C., McGrath, T., \& Nanda, N.
  (2023). Copy Suppression: Comprehensively Understanding a Motif in
  Language Model Attention Heads. arXiv:2310.04625
\item
  Meng, K., Bau, D., Mitchell, A., \& Yosinski, J. (2022). Locating and
  Editing Factual Associations in GPT. \emph{NeurIPS 2022}.
  arXiv:2202.05262
\item
  Park, K. et al.~(2023). The Linear Representation Hypothesis and the
  Geometry of Large Language Models. arXiv:2311.03658
\item
  Qwen Team (2025). Qwen2.5 Technical Report. arXiv:2412.15115
\item
  Rimsky, N., Gabrieli, N., Schulz, J., Tong, M., Hubinger, E., \&
  Turner, A. (2024). Steering Llama 2 via Contrastive Activation
  Addition. \emph{ACL 2024}. arXiv:2312.06681
\item
  Subramani, N. et al.~(2022). Extracting Latent Steering Vectors from
  Language Models. \emph{ACL 2022 Findings}. arXiv:2205.05124
\item
  Turner, A. et al.~(2023). Activation Addition: Steering Language
  Models Without Optimization. arXiv:2308.10248
\item
  Yona, G., Geva, M., \& Matias, Y. (2026). Hallucinations Undermine
  Trust; Metacognition is a Way Forward. \emph{ICML 2026 (Position)}.
  arXiv:2605.01428
\item
  Zou, A. et al.~(2023). Representation Engineering: A Top-Down Approach
  to AI Transparency. arXiv:2310.01405
\end{itemize}

\begin{center}\rule{0.5\linewidth}{0.5pt}\end{center}

\subsection{Appendix A: Experimental
Details}\label{appendix-a-experimental-details}

\subsubsection{A.1 Models}\label{a.1-models}

{\def\LTcaptype{none} 
\begin{longtable}[]{@{}lllll@{}}
\toprule\noalign{}
Model & Layers & Hidden dim & Vocab & Tied embed \\
\midrule\noalign{}
\endhead
\bottomrule\noalign{}
\endlastfoot
Gemma 2-2B-it (Gemma Team, 2024) & 26 & 2304 & \textasciitilde256k &
No \\
Gemma 2-9B-it (Gemma Team, 2024) & 42 & 3584 & \textasciitilde256k &
No \\
Llama 3.2-1B-Instruct (Dubey et al., 2024) & 16 & 2048 &
\textasciitilde128k & Yes \\
Qwen 2.5-1.5B-Instruct (Qwen Team, 2025) & 28 & 1536 &
\textasciitilde152k & Yes \\
Gemma 2-2B base (Gemma Team, 2024) & 26 & 2304 & \textasciitilde256k &
No \\
\end{longtable}
}

\subsubsection{A.2 Direction token sets}\label{a.2-direction-token-sets}

\textbf{Detection direction (hand-picked)}: \(\mathcal{U}\) = \{``I'',
``Unfortunately'', ``There'', ``It'', ``This''\}; \(\mathcal{C}\) =
\{``The'', ``In'', ``Paris'', ``Tokyo'', ``1''\}.

\textbf{Refusal direction}: \(\mathcal{R}\) = \{``No'', ``cannot'',
``doesn't'', ``I''\}; \(\mathcal{O}\) = \{``The'', ``Yes'', ``is'',
``It''\}.

These sets were selected to represent the first tokens of typical
uncertain/confident and refusing/complying model responses,
respectively. The sets are identical for all models (tokens are looked
up in each model's tokenizer). We tested 5 direction variants per
behavior; the separability contrast is invariant across all effective
variants.

\subsubsection{A.3 Fake entity
categories}\label{a.3-fake-entity-categories}

Capitals (Norlandia, Karvistan, Talmeris, \ldots), science (Pyrithanium,
Zorbane molecule, \ldots), culture (Zumareth Protocol, \ldots), dates
(Treaty of Valmora, \ldots), people (Dr.~Fenwick Hartley, \ldots). 10
per category, 50 total. The stress test (Section 6.3) extends to 60 fake
entities (30 Type 1 ``obvious'', 30 Type 2 ``subtle'') and 55 real
entities (25 easy, 20 obscure, 10 tricky-sounding).

\subsubsection{\texorpdfstring{A.4 Cross-model \(\alpha\)
calibration}{A.4 Cross-model \textbackslash alpha calibration}}\label{a.4-cross-model-alpha-calibration}

{\def\LTcaptype{none} 
\begin{longtable}[]{@{}
  >{\raggedright\arraybackslash}p{(\linewidth - 6\tabcolsep) * \real{0.1045}}
  >{\raggedright\arraybackslash}p{(\linewidth - 6\tabcolsep) * \real{0.2836}}
  >{\raggedright\arraybackslash}p{(\linewidth - 6\tabcolsep) * \real{0.4030}}
  >{\raggedright\arraybackslash}p{(\linewidth - 6\tabcolsep) * \real{0.2090}}@{}}
\toprule\noalign{}
\begin{minipage}[b]{\linewidth}\raggedright
Model
\end{minipage} & \begin{minipage}[b]{\linewidth}\raggedright
Effective \(\alpha\)
\end{minipage} & \begin{minipage}[b]{\linewidth}\raggedright
Residual norm (last layers)
\end{minipage} & \begin{minipage}[b]{\linewidth}\raggedright
\(\alpha\)/norm
\end{minipage} \\
\midrule\noalign{}
\endhead
\bottomrule\noalign{}
\endlastfoot
Gemma 2-2B & +15 & \textasciitilde70 & 0.21 \\
Llama 3.2-1B & −1 & \textasciitilde18 & 0.056 \\
Qwen 2.5-1.5B & +5 & \textasciitilde219 & 0.023 \\
\end{longtable}
}

The ratios span nearly 10×, indicating no simple linear scaling across
architectures.

\subsubsection{A.5 Signal decomposition
method}\label{a.5-signal-decomposition-method}

At each layer \(l\), we hook three positions: (1) residual
pre-attention, (2) mid-residual (after attention + residual add, before
feedforward), (3) residual post-feedforward. Attention contribution =
(2) − (1); MLP contribution = (3) − (2). Gemma 2's four layernorms per
layer (input, post-attention, pre-feedforward, post-feedforward) allow
clean decomposition.

\end{document}